\documentclass[letterpaper]{article} 
\usepackage{aaai24}  
\usepackage{times}  
\usepackage{helvet}  
\usepackage{courier}  
\usepackage[hyphens]{url}  
\usepackage{graphicx} 
\usepackage{natbib}  
\usepackage{caption} 
\usepackage{algorithm}
\usepackage{algorithmic}

\usepackage{breqn}
\usepackage{tabularx}
\usepackage{bibentry}
\usepackage{booktabs}
\usepackage[parfill]{parskip}
\usepackage{xcolor}
\newcommand{\TODO}{{\red \TODO}\xspace}

\usepackage{newfloat}
\usepackage{listings}
\DeclareCaptionStyle{ruled}{labelfont=normalfont,labelsep=colon,strut=off} 
\floatstyle{ruled}
\newfloat{listing}{tb}{lst}{}
\floatname{listing}{Listing}
\title{Leveraging Diffusion Perturbations for Measuring Fairness in Computer Vision}

\author{
    Nicholas Lui\textsuperscript{\rm 1}\equalcontrib,
    Bryan Chia\textsuperscript{\rm 1}\equalcontrib,
    William Berrios\textsuperscript{\rm 2},
    Candace Ross\textsuperscript{\rm 3},
    Douwe Kiela\textsuperscript{\rm 1,2}
}
\affiliations{
    \textsuperscript{\rm 1}Stanford University;
    \textsuperscript{\rm 2}Contextual AI;
    \textsuperscript{\rm 3}Meta AI\\
}

\begin{document}

\maketitle

\begin{abstract}
Computer vision models have been known to encode harmful biases, leading to the potentially unfair treatment of historically marginalized groups, such as people of color. However, there remains a lack of datasets balanced along demographic traits that can be used to evaluate the downstream fairness of these models. In this work, we demonstrate that diffusion models can be leveraged to create such a dataset. We first use a diffusion model to generate a large set of images depicting various occupations. Subsequently, each image is edited using inpainting to generate multiple variants, where each variant refers to a different perceived race. Using this dataset, we benchmark several vision-language models on a multi-class occupation classification task. We find that images generated with non-Caucasian labels have a significantly higher occupation misclassification rate than images generated with Caucasian labels, and that several misclassifications are suggestive of racial biases. We measure a model’s downstream fairness by computing the standard deviation in the probability of predicting the true occupation label across the different perceived identity groups. Using this fairness metric, we find significant disparities between the evaluated vision-and-language models. We hope that our work demonstrates the potential value of diffusion methods for fairness evaluations.
\end{abstract}

\section{Introduction}

Computer vision systems have been shown to replicate harmful statistical associations found in the training data~\cite{gebru, stock2018convnets,Wang_2019_ICCV}. Encoding such biases increases the risk of computer vision models unfairly treating under-represented groups~\cite{barocas2017problem}, such as people of color~\cite{gebru}. To combat this, there have been several efforts to mitigate bias, ranging from sampling~\cite{cao2020domain} to adversarial training~\cite{alvi2018turning}. However, there remains a lack of datasets, balanced along demographic traits, that are generally useful for evaluating the effectiveness of these techniques and the downstream fairness of computer vision models.

In this work, we explore the efficacy of diffusion methods~\cite{diffusionpaper, rombach2022high} for generating datasets balanced along demographic traits that can be used to evaluate the fairness of computer vision models. Our work is inspired by demographic perturbations in  language~\cite{maudslay2019s, smith2021hi, emmery2022cyberbullying, qian2022perturbation}.

We propose measuring fairness via \textbf{diffusion perturbations}, a novel diffusion-based approach that can be used to generate a dataset balanced along demographic traits, such as the perceived race of people. In our fully automated approach, we use diffusion models to generate a large set of base images. Then, each image is edited using inpainting to generate multiple variants, where each variant refers to a different demographic group. Image sets that do not meet the bar for realism and prompt fidelity are filtered using a combination of a VQA and face attribution model.

Images, in a given set of perturbations, share the same backgrounds and contexts, with the only difference being the perturbed demographic trait. To exploit the consistency within an image set, we propose a \textbf{fairness metric} that measures a model's robustness to demographic perturbations. Given a classification task, the fairness metric measures the extent to which the probability of the true label varies across different demographic groups. We draw inspiration from similar perturbation metrics in language \cite{ma2021dynaboard, qian2022perturbation, thrush2022dynatask}
and extend them to images.

Using diffusion perturbations, we construct a \textbf{novel dataset depicting different occupations balanced by perceived race}. Each occupation comprises 1,200 images per perceived identity group with 4 identity groups represented~(Black, Caucasian, East Asian, and Indian). We validate our dataset using crowdworkers, and show that our images demonstrate a high level of visual realism and a high probability of belonging to the intended target group.

Finally, we undertake a \textbf{fairness analysis using our occupations dataset}, where we evaluate several vision-language models (e.g. CLIP, FLAVA) using an occupation classification task. Using our fairness metric, we identify significant disparities between the models' robustness to demographic perturbations. We also find that images generated with non-Caucasian labels have a lower classification accuracy than Caucasian-labeled images, and that many of these misclassifications are suggestive of model biases.

\begin{figure*}[t]
   \centering
   \includegraphics[width=0.75\textwidth]{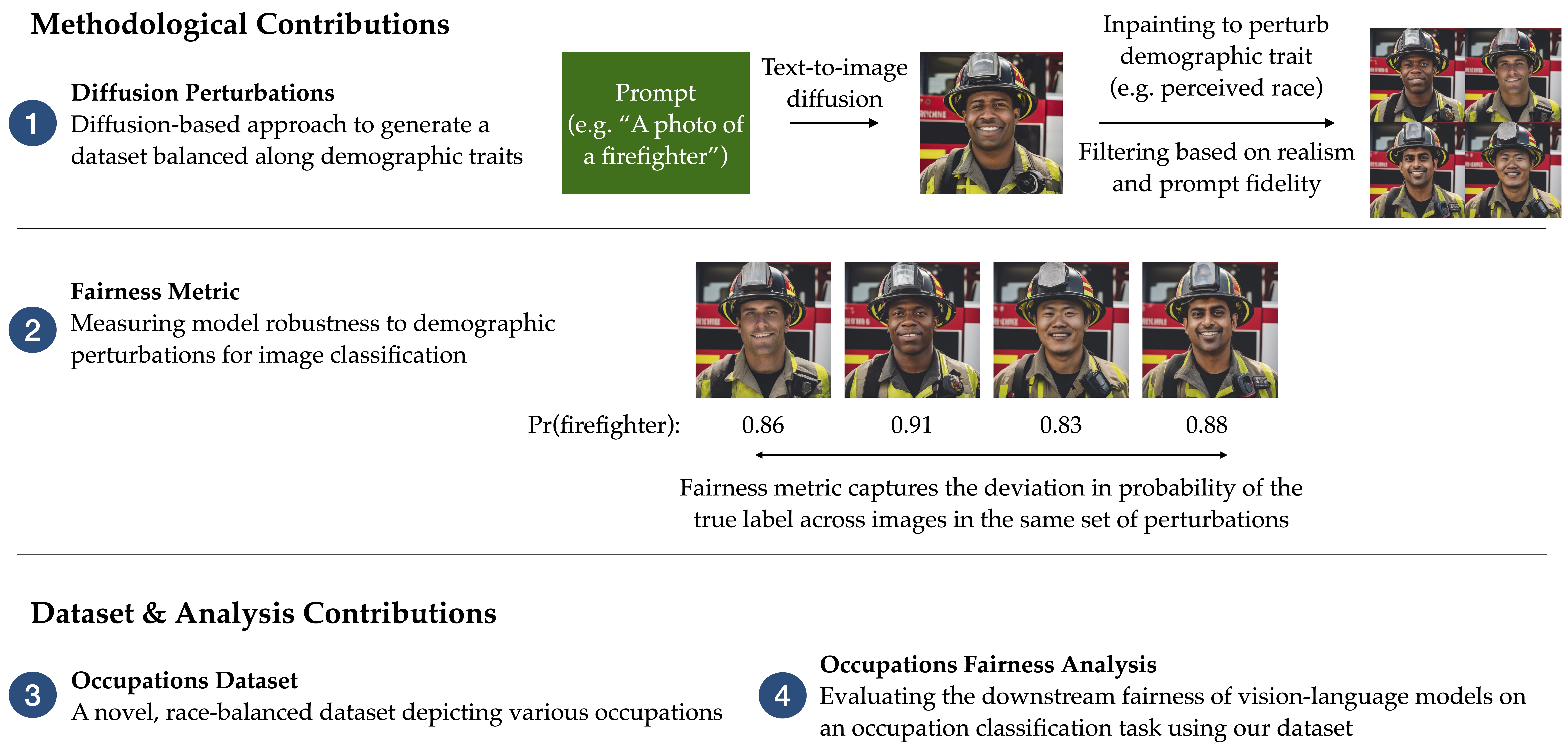}
   \caption{Our main contributions. We propose (1) a novel diffusion-based approach to generate a dataset balanced along demographic traits, and (2) a fairness metric to measure a model's robustness to demographic perturbations. We apply these techniques to (3) the creation of an occupations dataset and (4) produce fairness insights.}
   \label{fig:contributions}
\end{figure*}

The paper's main contributions are shown in Figure \ref{fig:contributions}. To enable greater exploration of our work, we release our generated dataset at this link: bit.ly/occupation-dataset. We release our code at this link: github.com/niclui/diffusion-perturbations.

\section{Related Work}

\subsubsection{Demographic perturbations of images.} To demographically perturb images, previous research has explored the use of generative adversarial networks (GANs)~\cite{yucer2020exploring, dash2022evaluating, jain2023zero}, as well as reinforcement learning-based approaches~\cite{wang2020mitigating}.

However, we favor a diffusion-based approach over a GAN-based one for two reasons. First, diffusion models have been shown to produce images with significantly more realism than GANs~\cite{dhariwal2021diffusion} and are thus a more suitable choice for producing realistic perturbations. Second, to achieve the balance between realism and faithfulness to the user input, GANs often require additional training data or loss functions for individual applications~\cite{meng2021sdedit}. In contrast, our diffusion-based approach does not require task-specific training to produce images that display a high level of realism and faithfulness.

\subsubsection{Datasets balanced along demographic traits.} 
Of the few datasets that are balanced along demographic traits, most of them comprise close-up face images with self-reported ethnicity and demographic information. One example is the FairFace dataset~\cite{karkkainen2021fairface}.

We believe that our diffusion perturbations approach extends these datasets in two key ways. First, by generating a set of perturbations which share the same backgrounds and contexts, we are able to isolate the correlation between the demographic trait and model predictions. We do not have this consistency across images in real-world datasets.

Second, while existing datasets are valuable for evaluating bias in facial recognition, their utility in other downstream tasks may be limited. With control over the input prompts and thus the content of images generated, text-conditioned diffusion models can generate datasets that are tailored for fairness evaluations on a wider array of downstream tasks. We demonstrate this capability by generating a race-balanced dataset for occupation classification.

Nonetheless, we acknowledge that existing datasets, with self-reported demographic information, are more likely to contain diverse representations of people from different demographic groups. In contrast, the training data that diffusion models are trained on may contain more limited representations of different demographic groups.

\section{Method}

In this section, we describe how we use text prompts to generate images, filter images to ensure high quality, and generate masks to perturb the perceived demographic trait.

\subsection{Prompt Creation}

The first step is compiling a list of prompts that we use to generate images. We use the following 5 occupations as they are distributed across white and blue collar jobs. For occupations that are more difficult to generate, we include the distinguishing attire of the occupation to improve identifiabilty. Our prompts begin with ``A photo of the face of":
\begin{enumerate}
\item{A car mechanic}
\item{A chef in a chef's jacket}
\item{A commercial pilot}
\item{A doctor in a white coat with a stethoscope}
\item{A firefighter}
\end{enumerate}

We opt to generate a set of base images rather than use real images of people for two reasons. First, inpainting for perturbations generally performs better on a base synthetic image. Second, using a mix of real and synthetic images could potentially confound downstream evaluations.

\begin{figure*}[t]
   \centering
   \includegraphics[width=0.76\textwidth]{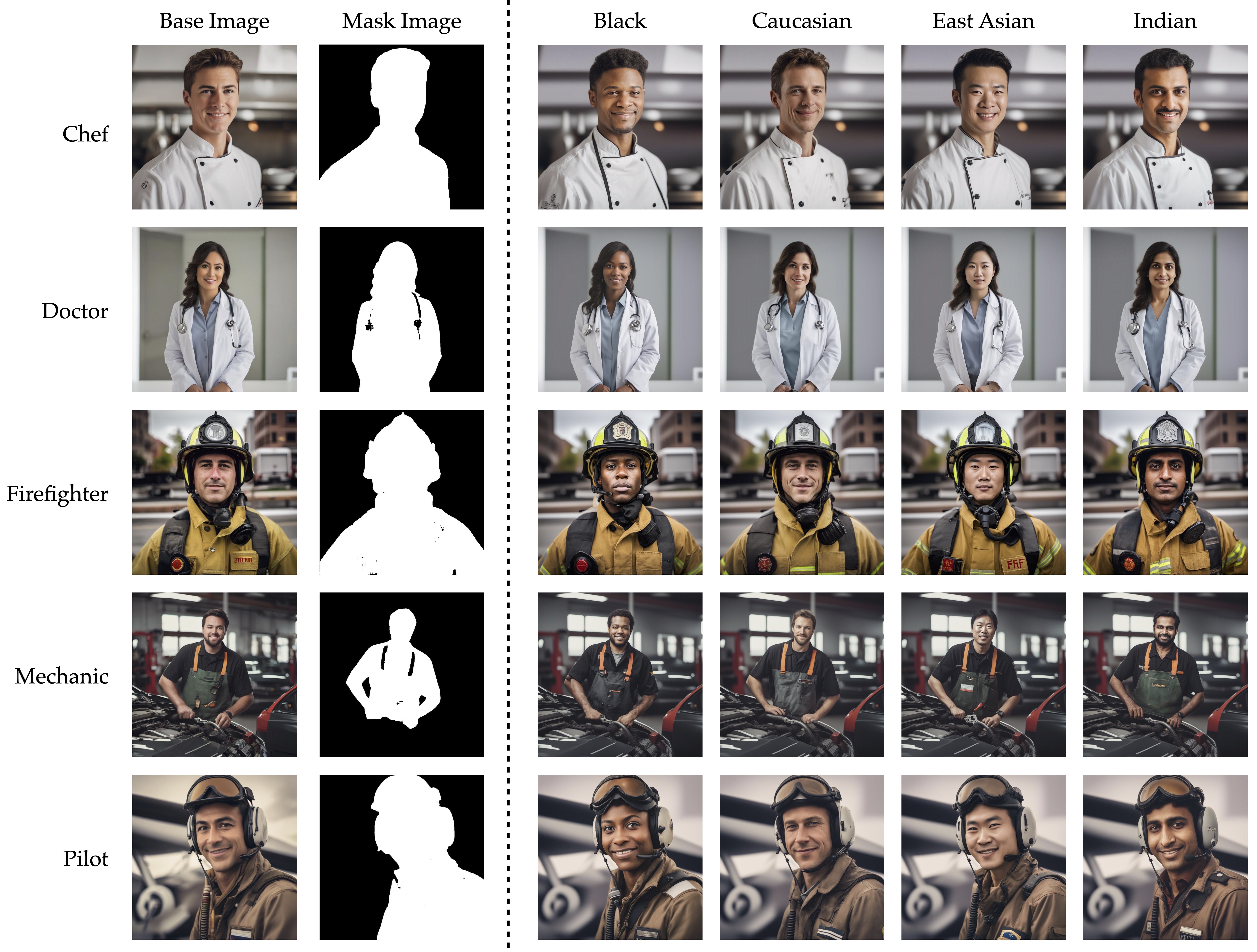}
   \caption{Samples of images generated for each occupation. The base image is generated using the text prompt ``A photo of the face of a $<$occupation$>$". We then generate a mask over the person(s) in the image. The original prompt is perturbed 4 times to include 4 different race identifiers: ``A photo of the face of a [Black$|$Caucasian$|$Asian$|$Indian] $<$occupation$>$". Base image-mask pairs are passed into the inpainting pipeline which produces four variants of the base image.}
   \label{fig:dataset_sample}
\end{figure*}

\subsection{Text-to-Image Generation}

We use the Stable Diffusion model~\cite{rombach2022high}, specifically Stable Diffusion XL for its improved photo-realism~\cite{podell2023sdxl}, to generate a 1024x1024 image for each prompt. In our initial experiments, we used an earlier version of Stable Diffusion (v2.1) with a grid search over a range of parameters to select the best image using a realism scorer. We find that images generated by Stable Diffusion XL, without any tuning, are significantly more realistic. We discuss our choice of hyperparameters in Appendix A1. For each occupation category, we generate 5-10k of images. We refer to these images as our base images.

\subsection{Automated Filtering of Base Image Using VQA Model}

We use a a ViLT-B/32 VQA model~\cite{kim2021vilt}, fine-tuned on VQAv2~\cite{goyal2017making}, to evaluate the base images. We evaluate the following:

\subsubsection{Text-to-image faithfulness.} Following~\citet{hu2023tifa}, we use the VQA model to evaluate the faithfulness of the generated image to its text input. Specifically, we ask the VQA model ``Is there a $<$occupation$>$ in this image?" [Q1] where $<$occupation$>$ is the occupation in the original prompt used to generate the image.

\subsubsection{Limb realism.} Diffusion models often face difficulties in producing realistic limbs. We thus ask the VQA model ``Are this person's limbs distorted?" [Q2].

\subsubsection{Overall realism.} The VQA model is asked: ``Is this image real or fake?" [Q3].

Our filtering proceeds in two stages. First, we only keep base images which answer ``Yes" to Q1 and ``No" to Q2. We record the yield for each occupation in Appendix A2. For each occupation, we select the top $\sim$2000 images with the highest score from the VQA model to Q3. From this set of $\sim$2000 images, we filter images that are grayscale by computing the number of unique colors in an image. This gives us a set of base images which we will perturb.

\subsection{Mask Generation}
We pass each base image through an end-to-end segmentation pipeline. The pipeline comprises two state-of-the-art models: First, the base image is passed through an open-set object detection model, Grounded-DINO~\cite{liu2023grounding}. Using the text prompt ``person", we obtain a bounding box around the person(s) in each image. Next, the base image and bounding box are passed into the Segment Anything model~\cite{kirillov2023segment} which identifies a segmentation mask while conditioning on the input box.

\subsection{Perturbation Using Inpainting}

We seek to perturb the perceived race of the people in our images. To do so, we use the Stable Diffusion inpainting pipeline~\cite{rombach2022high}. We pass a base image-mask pair into this pipeline, which inpaints the masked portion of the base image using the new prompt. The new prompt that we feed in is a perturbed version of the original prompt where we include a race identifier before the occupation (e.g. ``A photo of the face of a [Black$|$Caucasian$|$Asian$|$Indian]  firefighter"). The inpainting pipeline comprises 2 stages. First, the Stable Diffusion XL model~\cite{podell2023sdxl} is used to generate latents of the desired output size. Second, a specialized high-resolution refinement model~\cite{podell2023sdxl} applies the SDEdit image editing technique~\cite{meng2021sdedit} to the latents generated in the first step, producing a high-quality edited image.

We use four race categories - Caucasian, Black, East Asian, Indian. Three widely studied identity groups in the AI fairness literature are Caucasian, Black, and Asian. However, given text prompts including ``Asian", images generated by Stable Diffusion are typically ones we perceive as East Asian.\footnote{As such, for East Asian-labeled images, we use ``Asian" as the race identifier in the text prompt.} To introduce greater diversity in our dataset, we additionally include one of the largest Asian groups that is not East Asian (Indian). We note that some real-world face image datasets, such as UTKFace~\cite{zhifei2017cvpr}, have used these exact same race labels.

\subsection{Filtering Perturbed Sets Using FairFace}

The perturbation process is not guaranteed to produce realistic representations of different identity groups. To mitigate this risk, we use the FairFace model. It is a race attribution model trained on the FairFace dataset, which is a novel face image dataset that is balanced on race~\cite{karkkainen2021fairface}. The FairFace model exhibits significantly higher accuracy when applied to novel face image datasets compared to models trained on imbalanced race datasets. Moreover, it maintains consistent accuracy across various race and gender groups. We use the 4-race version of the model which uses the exact same labels that we do (``Black", ``Caucasian", ``East Asian", ``Indian"). Using the FairFace model, we keep only sets of perturbed images where all 4 images are classified to the intended category. To maintain an even distribution, we sample 1200 image sets from each occupation. Our dataset comprises 24k images (1200$\times$4 perceived races$\times$5 occupations), with example images in Figure \ref{fig:dataset_sample}.

\subsection{Limitations of Image Generation}
\label{ssec:limitations}

Our approach yields edited images that are high quality and consistent with the base image. However, we note the following challenges: First, while we seek to generate images that are race-balanced, there is no guarantee that our images are balanced along other demographic traits. For instance, individuals in our base and edited images could be perceived as being more masculine. This could be attributed to bias in Stable Diffusion as many of our chosen occupations have a male skew~\cite{luccioni2023stable}. This issue limits the generalizability of our analysis to examining racial disparities within certain demographic boundaries (e.g. individuals who are perceived to be more masculine). To mitigate concerns about gender bias, we perform a robustness check using a regenerated sample of the dataset where we specify the perceived gender in each prompt (see Section \ref{ssec:results_difficult}).

Second, for occupations with a white skew in the US (e.g. pilot), we find that the base image tends to be Caucasian-presenting, possibly due to bias in Stable Diffusion. One question is whether the perceived race of the base image affects the quality of perturbations for other identity groups. However, since we use diffusion inpainting~\cite{rombach2022high, wang2023imagen}, the individual in the base image is masked and it is less likely that their perceived race will have downstream effects on the perturbation quality. This belief is validated through our crowdworker review, which shows that there is no statistically significant difference in the image quality between the perceived identity groups.

\subsection{Dataset Review}

To validate the efficacy of the aforementioned pipeline, we conducted an Amazon MTurk survey over an evenly distributed sample of 4000 images ($16.7\%$ of the dataset). We asked workers two questions: (1) \textit{``Does this image contain obvious quality issues with the person? (e.g. blurred out facial features, additional limbs)"}, with ``Yes", ``No", and ``Unsure" as possible responses, and (2) \textit{``What is one identity group that this person is likely to belong to?"} with ``Black'', ``Caucasian", ``East Asian (e.g. Chinese)", ``South Asian (e.g. Indian)", and ``Others" as possible responses. We present our results in Table \ref{tab:review} and discuss further in the Appendix A3, where we show that the results are similar across the different perceived races. 

\begin{center}
\captionof{table}{Human dataset review: ``Realism Score" is the proportion of reviewers who did not find quality issues with the image; ``Race Fidelity Score" is the proportion of reviewers who indicated that the perceived race of the person is the same as the race specified during image editing.}

\begin{tabular}{lrr}
  \toprule
      Data & Realism Score & Race Fidelity Score\\
      \midrule
     \textbf{Overall} & \textbf{85.1\%}  & \textbf{91.0\%}\\
     Chef & 84.9\% & 90.6\%\\
     Doctor & 86.6\% & 86.9\%\\
     Firefighter & 91.5\% & 92.1\%\\
     Mechanic & 77.4\% & 95.5\%\\
     Pilot & 85.0\% & 90.0\%\\
  \bottomrule
 \end{tabular}
 \label{tab:review}
\end{center}

\section{Fairness Task}

Our downstream task is multi-class occupation classification where the model chooses from a set of occupation labels. Our analysis does not try to show that one model is generally more or less fair than another on occupation classification. Instead, we are evaluating the relative fairness of models for a specific set of occupations with a specific set of labels.  We consider two label sets:

\subsubsection{Base label set.} We use the same set of labels for all occupations. The label set contains each original label as well as a negative label that is similar but clearly distinct from the original occupation. Labels used: [``chef", ``server", ``doctor", ``nurse", ``pilot", ``driver", ``mechanic", ``engineer", ``firefighter", ``police officer"].

\subsubsection{Difficult label set.} To provide a more challenging benchmark, we consider a more difficult set of occupation-specific labels as shown in Table \ref{tab:labels_table}. For each occupation, the label set includes the true occupation and 7 other adjacent occupations that we selected based on occupations listed in the U.S. Bureau of Labor Statistics. We include negative labels that are in the same line of work as the true label but with differing responsibilities (e.g. ``doctor" and ``nurse").

A few of our negative labels may not be easily distinguishable in appearance from the true label (e.g. ``doctor" and ``physician assistant"). That being said, if the model is truly fair, we would expect it to choose the true label over the contending one with the same probability for all perceived identity groups.

To create a label set more appropriate for models trained on a larger amount of data, we use occupations that are less well-known, and thus better suited for models that have encountered them in their larger training datasets.

\begin{center}
 \captionof{table}{Difficult label set for multi-class classification.}
 \begin{tabular}{ll}
  \toprule
      Occupation & Adjacent Occupations\\
      \midrule
      Chef 
      & Line cook, Cafeteria attendant,  Waiter,\\
      & Dishwasher,  Food preparation worker \\
      & Host, Server\\
      \midrule
     Doctor 
     & Nurse, Physician assistant, Veterinarian,\\ 
     & Clinical laboratory technician, Pharmacist,\\ 
     & Emergency medical technician, Midwife\\ 
     \midrule
     Firefighter & Fire chief, Coast guard, Security guard,\\
     & Paramedic, Pilot, Police officer, Soldier \\
     \midrule
     Mechanic & Automobile engineer, Civil engineer,\\
     &Aerospace engineer, Mechanical engineer,\\
     &Electrical engineer, Industrial engineer,\\
     & Petroleum engineer \\
     \midrule
     Pilot & Flight steward, Flight stewardess, Driver  \\ 
     &  Aircraft fueler, Airline reservation agent,  \\
     & Air traffic controller, Aircraft engineer\\
  \bottomrule
 \end{tabular}
\label{tab:labels_table}
\end{center}

\subsection{Models Evaluated}

Table \ref{tab:models_eval} lists the evaluated models. We evaluate FLAVA~\cite{singh2022flava} and unless otherwise stated, the ViT-B/32 variant of CLIP~\cite{radford2021learning}. For all models, we compute the cosine similarity between the image and each possible label which is prepended by ``A photo of''. The set of cosine similarities is passed into a softmax function to generate a set of prediction probabilities.

\begin{center}
\captionof{table}{Models evaluated on occupation classification task.}
\begin{tabular}{llr}
  \toprule
      & Training & Dataset\\
      Model & Dataset & Size\\\midrule
     FLAVA & PMD  & 70M\\
     CLIP-OpenAI & WebImageText & 400M\\
     CLIP-LAION400M & Common Crawl & 400M\\
     CLIP-LAION2B & Common Crawl & 2B\\
  \bottomrule
 \end{tabular}
\label{tab:models_eval}
\end{center}

\subsection{Fairness Metric}

We define an extrinsic fairness metric to measure robustness to demographic perturbation in our classification task. If a model is fair, perturbing the demographic group should have minimal effect on model performance. We thus construct a fairness metric that captures the standard deviation in probability of the true label within an image set.

Formally, we have a classifier $f_C$ and a dataset $X$. The dataset $X$ comprises $N$ image sets and each image set contains $K$ perturbed images. For image $j$ from set $i$, our classifier outputs the probability of the true label, $f_C(x_{ij})$. We compute the standard deviation of this probability across all $j$ images to give us the standard deviation for set $i$. We then take the median standard deviation across all $N$ sets.

Our fairness metric is 1 minus the median standard deviation. A fairness metric close to 1 (i.e. median standard deviation that is close to 0) implies that the model does equally well on every perceived identity group.

\begin{dmath*}
 F_M(f_C, X) =
 1 - med
 \{
 \sqrt{
 \frac{
 \sum_{j=1}^{K} (f_C(x_{ij}) - \overline{f_C(x_{ij})}
 )^{2}} {K-1}
 } 
 \}
 _{i=1,...,N}
\label{eq:fair_metric}
\end{dmath*}

\section{Results}

\subsection{Results for Base Labels}

Table \ref{tab:easy_fairness} reports the results of the fairness metric across the different models. The models all achieve a very high level of parity across the different perceived identity groups. This necessitates a much harder set of labels to benchmark the models' downstream fairness.

The table also shows the classification accuracy rate. However, we are primarily interested in the fairness metric. Regardless of what the average accuracy is, the model should be doing equally good/bad across the different perceived identity groups.

\begin{center}
 \captionof{table}{Models evaluated on our downstream occupation classification task using base label set.}
 \begin{tabular}{lrr}
  \toprule
 & \textbf{Fairness} & Classification\\ 
 Model & \textbf{Metric} &  Accuracy\\\midrule
  FLAVA & 0.990 & 92.0\% \\ 
  CLIP-LAION400M & 0.997 & 98.3\% \\
  CLIP-LAION2B& \textbf{0.998} & \textbf{98.9\%} \\
  CLIP-OpenAI & 0.983 & 95.2\% \\ 
  \bottomrule
 \end{tabular}
\label{tab:easy_fairness}
\end{center}

\subsection{Results for Difficult Labels}
\label{ssec:results_difficult}

Table \ref{tab:hard_fairness} reports the results of the fairness metric across the different models. We perform pairwise comparisons for the fairness metric using Mood's non-parametric median test and Bonferroni's Correction to account for multiple hypothesis testing. We find that FLAVA has the highest fairness metric at the 1\% significance level. Within the CLIP models, we find that CLIP-OpenAI $>$ CLIP-LAION2B $>$ CLIP-LAION400M at the 1\% level. The table also shows the accuracy rates for the different models.

\begin{center}
\captionof{table}{Models evaluated on our downstream occupation classification task using difficult label set.}
 \begin{tabular}{lrr}
  \toprule
    & \textbf{Fairness} & Classification\\ 
    Model & \textbf{Metric} &  Accuracy\\\midrule
    FLAVA & \textbf{0.983}  & \textbf{90.1\%} \\
    CLIP-LAION400M & 0.835 & 75.1\% \\
    CLIP-LAION2B& 0.849 & 79.8\%\\
    CLIP-OpenAI & 0.884 & 68.6\% \\
  \bottomrule
\end{tabular}
\label{tab:hard_fairness}
\end{center}

Although FLAVA has the smallest dataset size (trained on $>$5x less data than the CLIP models), it has the highest accuracy. Surprisingly, FLAVA achieves a similar accuracy on the difficult labels as it does on the base labels, while the other CLIP models all see a large decrease. FLAVA switches from being the least accurate model on the base labels to being the most accurate model on the difficult labels.

We hypothesize that because FLAVA is trained on a much smaller dataset than CLIP, it has a more limited vocabulary. Consequently, it might not be attempting to discern between a ``doctor" vs ``physician's assistant" for instance, and is thus not tricked by the difficult label set. Thus, FLAVA outperforms all the CLIP models in both accuracy and the fairness metric. While CLIP's richer model embedding space allows us to use more granular labels, it might also make it easier to draw out biases from the model.

However, this risk can be mitigated if the larger dataset used to develop richer embeddings is well curated. Within the CLIP models, CLIP-OpenAI scores the highest fairness metric. It achieves a significantly higher metric than CLIP-LAION400M, despite the two models having a similar architecture and training dataset size. The different dataset used could be responsible for disparities in the fairness metric.~\citet{birhane2021multimodal} found that there were major data filtering issues with LAION-400M, resulting in the dataset containing large amounts of racist content and stereotypes. Thus, CLIP-LAION400M may have encoded racial biases to a larger extent than CLIP-OpenAI which was likely trained on a better curated dataset.

CLIP-LAION2B also has a lower fairness metric than CLIP-OpenAI, despite training on roughly 5x the amount of data. This could suggest that there are diminishing returns to fairness improvements with dataset size, especially if the underlying data contains significant biases. 

\subsubsection{Dataset robustness.} In our dataset curation, we used only base images that a VQA model assessed to have included the target occupation. However, this might introduce biases in our evaluations in favor of models trained on a similar dataset as the VQA model. Thus, we repeat the base experiment on a subset of the data where the base image was correctly classified by all the evaluated models. We have the same fairness metric ordering: FLAVA (0.98), CLIP-OpenAI (0.884), CLIP-LAION2B (0.861), CLIP-LAION400M (0.835). Full results are provided in the Appendix A4.

\subsubsection{Label robustness.} For a given occupation, we use 7 negative labels. However, some labels may be simply adding noise to the results. We remove every negative label except the top misclassified label and see if the results still hold. We retain the same ordering in the fairness metric with this experiment: FLAVA (0.999), CLIP-OpenAI (0.876), CLIP-LAION2B (0.866), CLIP-LAION400M (0.847). Full results are provided in the Appendix A5.

\subsubsection{Ablation study on number of parameters.} We use the ViT-B/32 variant of CLIP, but it's not clear if our results hold for a larger parameter size. We evaluate the ViT-L/14 variant of CLIP, which has close to 3X the number of parameters. We find that CLIP-LAION400M (ViT-L/14) still has the lowest fairness metric, while CLIP-OpenAI (ViT-L/14) and CLIP-LAION2B (ViT-L/14) have similar fairness metrics. Our results are in Table \ref{tab:ablation}. We continue to use the ViT-B/32 variant for our analysis.

\begin{center}
 \captionof{table}{Ablation Study. We evaluate the ViT-L/14 variant of CLIP on our downstream occupation classification task.}
 \begin{tabular}{lrr}
  \toprule
 & \textbf{Fairness} & Classification\\ 
 Model & \textbf{Metric} &  Accuracy\\\midrule
  CLIP-LAION400M & 0.862 & 82.4\% \\
  CLIP-LAION2B& \textbf{0.908} & \textbf{90.1\%} \\
  CLIP-OpenAI & 0.903 & 84.1\% \\ 
  \bottomrule
 \end{tabular}
 \label{tab:ablation}
\end{center}

\subsubsection{Robustness to gender bias in image generation and perturbation.} In Section \ref{ssec:limitations}, we note that the occupations we evaluate may have a male skew, resulting in generated images that are largely male-presenting. In our error analysis (Appendix A9), we also find that the presenting gender in the inpainted image may sometimes differ from the original base image.

To ensure that our results are robust to these biases, we regenerate a small sample of our dataset, while ensuring that it is balanced across both the male and female genders. We do so by specifying a gender in the prompt used for image generation and inpainting (e.g. ``A photo of the face of a [male$|$female] firefighter"). Given that we specify the same gender in the prompts used for base image generation and inpainting, it is much less likely that the presenting gender in the inpainted image will differ from that of the original base image. For each occupation, our sample comprises 200 images (100 male, 100 female) for each perceived race.

Using this sample dataset, we find that our fairness metric ordering is retained (results in parentheses): FLAVA (0.944) $>$ CLIP-OpenAI (0.874) $>$ CLIP-LAION2B (0.827) $>$ CLIP-LAION400M (0.821). These results suggest that our qualitative results are robust to gender bias in image generation and perturbation.

\subsection{Occupation-Level Analyses.}

Across the board, all 3 non-Caucasian perceived identity groups have a lower classification accuracy than Caucasians at the 1\% level (Black: -6.09\%, East Asian: -3.21\%, Indian: -3.20\%). Table \ref{tab:occ_hard_fairness} shows the results of the fairness metric for each occupation. Within CLIP models, CLIP-LAION2B achieves the highest fairness metric on Chef, while CLIP-OpenAI does the best on every other occupation.

\begin{center}
\captionof{table}{Fairness Metric by Occupation.
\\Doc: Doctor, FF: Firefighter, Mec: Mechanic}
\begin{tabular}{lrrrrr}
 \toprule
Model & Chef & Doc & FF & Mec & Pilot \\\midrule
FLAVA & \textbf{0.999}  & 0.728 & \textbf{0.988} & \textbf{0.986} & \textbf{0.973} \\\midrule
 CLIP &  0.837 & 0.830 & 0.946 & 0.791 & 0.747 \\
 LAION400M & & & & & \\\midrule
 CLIP & 0.956 & 0.740 &  0.931 & 0.798 & 0.749 \\
 LAION2B & & & & & \\\midrule
CLIP &  0.895 & \textbf{0.867} & 0.955 & 0.816 & 0.859 \\
OpenAI & & & & & \\
 \bottomrule
\end{tabular}
\label{tab:occ_hard_fairness}
\end{center}

In the Appendix A7, we illustrate how different models perform when predicting the occupation of the image across different perceived identity groups. To get race-level effects, we regress a binary outcome variable (1 if image predicted correctly, 0 otherwise) on perceived race, using robust standard errors clustered at the image set level to account for within-set correlation.

We show findings from 3 occupations below, but analyze all 5 occupations in the Appendix A7.

\textbf{Chef:} CLIP-LAION400M and CLIP-OpenAI predict a Black chef with an accuracy 10-13\% lower than Caucasians. CLIP-LAION400M predicts an East Asian and Indian Chef with a 10\% and 9\% lower accuracy respectively. This disparity can be explained by CLIP-LAION400M and CLIP-OpenAI predicting non-Caucasian chefs as line cooks more often than Caucasians. Summing across both models, Black chefs are predicted as line cooks 104\% more times than Caucasians. East Asians and Indians are predicted as line cooks 72\% and 57\% more times respectively.

\textbf{Doctor:} The top misclassified label is physician assistant. This is understandable as a physician assistant and doctor has similar appearances. Still, it does not nullify the fact that we should see an equal probability of predicting either class for the different perceived identity groups. Instead, we find a significantly lower difference in predicting a Black doctor than Caucasians (ranging from $10-24\%$) across the CLIP-based models. Black doctors are misclassified as physician assistants 34\% more times than Caucasians. For CLIP-LAION400M, Indians and East Asians have a $12\%$ \textit{higher} accuracy than Caucasians. In a cursory probe, we find that there is a corresponding pattern in classifying Caucasian-presenting doctors within CLIP-LAION2B's training set as doctors more often than Black-presenting doctors, who are classified as physician assistants. We provide further details in Appendix A8.

\textbf{Pilot:}  We find that the probability of CLIP models predicting Black, Indian, and East Asian pilots are significantly lower as compared to Caucasian pilots, with most of these models predicting ``flight stewards" instead. Given that $95.7\%$ of employed pilots are Caucasian (according to the US Bureau of Labor Statistics), it is possible that there is a lack of diversity in representations of pilot in the US-centric portions of the training data.

\subsubsection{Error analysis.} For every non-Caucasian perceived identity group, we sample 10 image sets where the Caucasian-labeled image is correctly classified and the non-Caucasian group is not. We present the images for each occupation in the Appendix A9. For most sets, we find that the misclassified images look similar to the correctly classified Caucasian-labeled image, suggesting that it may be model bias, rather than image quality, that is driving inequitable model performance.

\subsection{Implicit Association Test (IAT)}
Drawing from the racial IAT~\cite{greenwald1998measuring, maina2018decade} to analyze subvert biases, we probe the model to choose one of two labels: ``A trustworthy/untrustworthy person". We report results in Appendix A6. We find that CLIP-LAION400M has the largest spread in probabilities across the perceived race groups, consistent with our finding that CLIP-LAION400M could be less fair than the other two CLIP models in some situations.

\subsection{Evaluation With Large Language Models (LLMs)}

Finally, we are interested in assessing the fairness of LLMs using difficult labels in our dataset. For this, we employed LENS~\cite{berrios2023language}, a model that reasons and performs classification tasks based on visual descriptions. We use the prompt "Question: Can you please identify the occupation that best represents the image? Short Answer: \{answer\}" to select the best continuation. Since LENS provides the log probabilities for each possible continuation, we calculate the joint probability and then employ the same calculation as in CLIP or FLAVA.

\pagebreak

In our case, LENS (H\textsubscript{14}-FlanT5\textsubscript{xxl}) achieved a fairness metric of 0.837, while LENS (H\textsubscript{14}-FlanT5\textsubscript{xl}) reached 0.965. These results suggest that LENS (H\textsubscript{14}-FlanT5\textsubscript{xl}) may be performing more fairly on our task than the CLIP models. Finally, Table \ref{tab:lensExp_occ_hard_fairness} shows occupation-specific fairness metric results across both LENS models.

\begin{center}
\footnotesize
\captionof{table}{Fairness Metric by Occupation for LENS models.
\\Doc: Doctor, FF: Firefighter, Mec: Mechanic}
\begin{tabularx}{\columnwidth}{lrrrrr}
 \toprule
Model (LENS) & Chef & Doc & FF & Mec & Pilot \\\midrule
  H\textsubscript{14}-FlanT5\textsubscript{xxl} & 0.928  & 0.826  & 0.824 & 0.762 & 0.916 \\\midrule
 H\textsubscript{14}-FlanT5\textsubscript{xl} &0.984 & 0.894 & 0.978 & 0.939 & 0.986 \\
 \bottomrule
\end{tabularx}
\label{tab:lensExp_occ_hard_fairness}
\end{center}

\section{Conclusion}

As computer vision models grow larger and more powerful, there is an increasing need to develop robust techniques to audit their fairness on downstream tasks. We propose a novel diffusion-based approach to generate datasets balanced along demographic traits that can be used to assess relative model fairness. In our approach, a large set of images is generated using diffusion models and then perturbed to generate multiple variants representing different demographic groups. We develop a dataset on occupation classification, and show that our approach can be used to expose model biases. We also develop a fairness metric to measure a model's robustness to demographic perturbations. Using our fairness metric and generated dataset, we uncover disparities between several vision-language models. We hope that our work demonstrates the value of diffusion models to fairness evaluations and research.

\section{Ethical Statement}
While our approach shows promise in building a dataset balanced along demographic traits for fairness evaluations, we recognize that there are key ethical considerations that need to be addressed.

First, we acknowledge that race is socially constructed and that it is impossible to identify someone's race solely based on their appearance. Racial identity is a personal decision and there is no fixed definition of what people of a specific race should look like. Nonetheless, we recognize that a lot of the harms from racism come from racial profiling~\cite{glover2009racial}, which is heavily influenced by a person's appearance and the extent to which the individual has physical characteristics that societal perceptions associate with their race~\cite{maddox2018racial}. Despite the limitations of Stable Diffusion, we believe that the generated images contain some of those characteristics. Nonetheless, we acknowledge that our work may have the unintended consequence of reifying a fixed definition of race.

Second, diffusion models may not generate images that reflect the diversity of people belonging to different identity groups. Image generations build on the representations of different identity groups found in the training dataset, which may be especially narrowly defined for people of color. In future work, we hope to evaluate methods like FairDiffusion~\cite{friedrich2023fair} which can instruct generative models to produce images that are fairer and more diverse. We discuss other limitations of our image generation process in Section \ref{ssec:limitations}.

\section{Resources}

The Appendix can be found at this link: bit.ly/dp-appendix. To enable greater exploration and improvement on our work, we release our dataset here: bit.ly/occupation-dataset. We release our code at this link: github.com/niclui/diffusion-perturbations.

\section{Acknowledgements}

We would like to thank Helen Gu, Amir Hertz, Gautam Mittal, Rajan Vivek, Rohan Koodli, and Melissa Hall for their thoughtful feedback. This work was funded in part by a cloud compute grant from Google and the Stanford Institute for Human Centered Artificial Intelligence. NL is supported by a Stanford Knight-Hennessy Fellowship. Meta was not involved in running any models for the dataset creation or evaluation.

\bibliography{aaai24}

\end{document}